%% file: main.tex
\documentclass{article} % For LaTeX2e

\input{math_commands.tex}

\usepackage{hyperref}
\usepackage{url}

\usepackage[accepted]{icml2025}

\usepackage[utf8]{inputenc} % allow utf-8 input
\usepackage[T1]{fontenc}    % use 8-bit T1 fonts
\usepackage{hyperref}       % hyperlinks
\usepackage{url}            % simple URL typesetting
\usepackage{booktabs}       % professional-quality tables
\usepackage{amsfonts}       % blackboard math symbols
\usepackage{nicefrac}       % compact symbols for 1/2, etc.
\usepackage{microtype}      % microtypography
\usepackage{xcolor}         % colors

\usepackage{amsthm}
\usepackage{algorithm,algorithmic}
\renewcommand{\algorithmiccomment}[1]{\bgroup\hfill//~#1\egroup}
\usepackage{amsmath}
\usepackage{amssymb}
\usepackage{mathtools}
\usepackage{enumitem}
\setlist[itemize]{noitemsep, topsep=0pt, leftmargin=11pt}
\setlist[enumerate]{noitemsep, topsep=0pt, leftmargin=11pt}
\usepackage{wrapfig}
\usepackage{xcolor}
\usepackage{subcaption}
\usepackage{soul}
\usepackage{multirow}

\usepackage[capitalize,noabbrev]{cleveref}

\theoremstyle{plain}
\newtheorem{theorem}{Theorem}[section]

\theoremstyle{definition}
\newtheorem{definition}[theorem]{Definition}

\theoremstyle{remark}

\def\1{\mathbf{1}}
\def\P{\mathbb{P}}

\def\N{\mathbb{N}}

\newcommand{\mc}[1]{\mathcal{#1}}

\usepackage[textsize=tiny]{todonotes}

\icmltitlerunning{GPT-4o as the Gold Standard: A Scalable and General Purpose Approach to Filter Language Model Pretraining Data}

\begin{document}

\twocolumn[
\icmltitle{GPT-4o as the Gold Standard: A Scalable and General Purpose Approach to Filter Language Model Pretraining Data}




\begin{icmlauthorlist}
\icmlauthor{Jifan Zhang}{wisc}
\icmlauthor{Ziyue Luo}{osu}
\icmlauthor{Jia Liu}{osu}
\icmlauthor{Ness Shroff}{osu}
\icmlauthor{Robert Nowak}{wisc}
\end{icmlauthorlist}

\icmlaffiliation{wisc}{University of Wisconsin-Madison}
\icmlaffiliation{osu}{Ohio State University, Columbus}

\icmlcorrespondingauthor{Jifan Zhang}{jifan@cs.wisc.edu}

\icmlkeywords{Large Language Model Pretraining, Data Filtering, Active Learning}

\vskip 0.3in
]



\printAffiliationsAndNotice{}  % leave blank if no need to mention equal contribution

\begin{abstract}
Large language models require vast amounts of high-quality training data, but effective filtering of web-scale datasets remains a significant challenge. This paper demonstrates that GPT-4o is remarkably effective at identifying high-quality training data, but its prohibitive cost makes it impractical at web-scale. We propose SIEVE, a lightweight alternative that matches GPT-4o accuracy at less than 1\% of the cost. SIEVE can perform up to 500 filtering operations for the cost of one GPT-4o filtering call. The key to SIEVE is a seamless integration of GPT-4o and lightweight text classification models, using active learning to fine-tune these models in the background with a small number of calls to GPT-4o. Once trained, it performs as well as GPT-4o at a tiny fraction of the cost. Through different filtering prompts, SIEVE can efficiently curate high quality data for general or specialized domains from web-scale corpora -- a valuable capability given the current scarcity of high-quality domain-specific datasets. Extensive experiments using automatic and human evaluation metrics show that SIEVE and GPT-4o achieve similar performance on five highly specific filtering prompts. In addition, when performing quality filtering on web crawl datasets, we demonstrate SIEVE can further improve over state-of-the-art quality filtering methods in the DataComp-LM challenge for selecting LLM pretraining data.
\end{abstract}

\section{Introduction}
Large Language Models (LLMs) have revolutionized natural language processing, demonstrating remarkable capabilities across a wide range of tasks. As the field progresses, there is growing interest in both developing specialized LLMs tailored to specific domains and improving the quality of training data for general-purpose LLMs~\citep{lee2023benefits,gupta2024rag,li2024small}. A critical component in both scenarios is the curation of high-quality datasets, whether domain-specific or general-purpose, for training and fine-tuning these models. However, the task of filtering vast amounts of web-scale data presents significant challenges in terms of cost, time, and effectiveness.

\begin{figure*}[t]
    \centering
    \includegraphics[width=.8\linewidth]{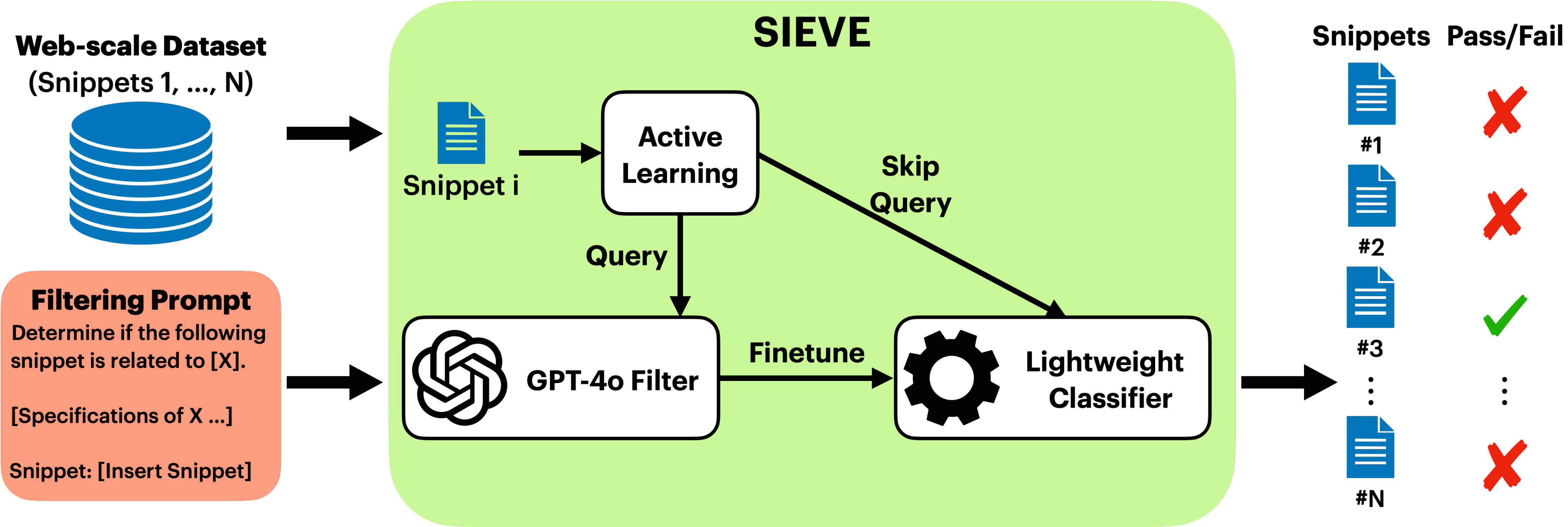}
    \caption{\textbf{System Overview.} From user's perspective, SIEVE acts as if applying GPT-4o with the filtering prompt to all text snippets in a web-scale dataset. The output from the SIEVE system is the set of all text snippets that receive a `pass'. To reduce the prohibitively high cost of applying GPT-4o on every snippet, SIEVE utilizes active learning to distill lightweight filtering models based on pretrained encoders (e.g., T5 or DeBERTa), effectively reducing the overall cost to less than $1\%$.}
    \label{fig:sieve_system}
    \vspace{-\intextsep}
\end{figure*}
Existing approaches to data filtering and curation primarily rely on two strategies: (1) sourcing data from specific, trusted sources, and (2) employing pre-existing quality and toxicity detection models to filter harmful content. For instance, medical LLMs often utilize datasets derived from PubMed to ensure domain specificity, while general-purpose LLMs typically employ automated quality metrics to select high-quality samples from web crawls. While these data filtering methods have their merits, they suffer from notable limitations in terms of both flexibility and comprehensiveness. Many domains lack comprehensive, exclusive sources of high-quality text data, and pre-trained detection models typically focus on general quality metrics and toxicity, limiting their applicability for domain-specific queries or customized quality rubrics. Furthermore, relying solely on established data sources ignores the vast amount of potentially valuable information available on the broader internet, leading to datasets that may be limited in scope and diversity. These constraints can significantly hinder both the development of specialized language models and the efficiency of general-purpose model training.

Recent advancements in general-purpose language models, such as GPT-4o, offer a potential solution to these challenges. These models can act as effective filtering mechanisms when provided with appropriate filtering prompts, whether for identifying domain-specific content or assessing general text quality. To illustrate as an example, one could employ GPT-4o to iterate over all text snippets of the internet to determine whether each piece pertains to the 2024 presidential election or meets specific quality criteria for LLM pretraining. In this context, the machine learning model serves as a binary classification mechanism, evaluating one snippet at a time. However, the computational cost of applying models like GPT-4o to web-scale datasets is prohibitively expensive for most organizations.

\begin{figure*}[ht!]
    \centering
    \includegraphics[width=.7\linewidth]{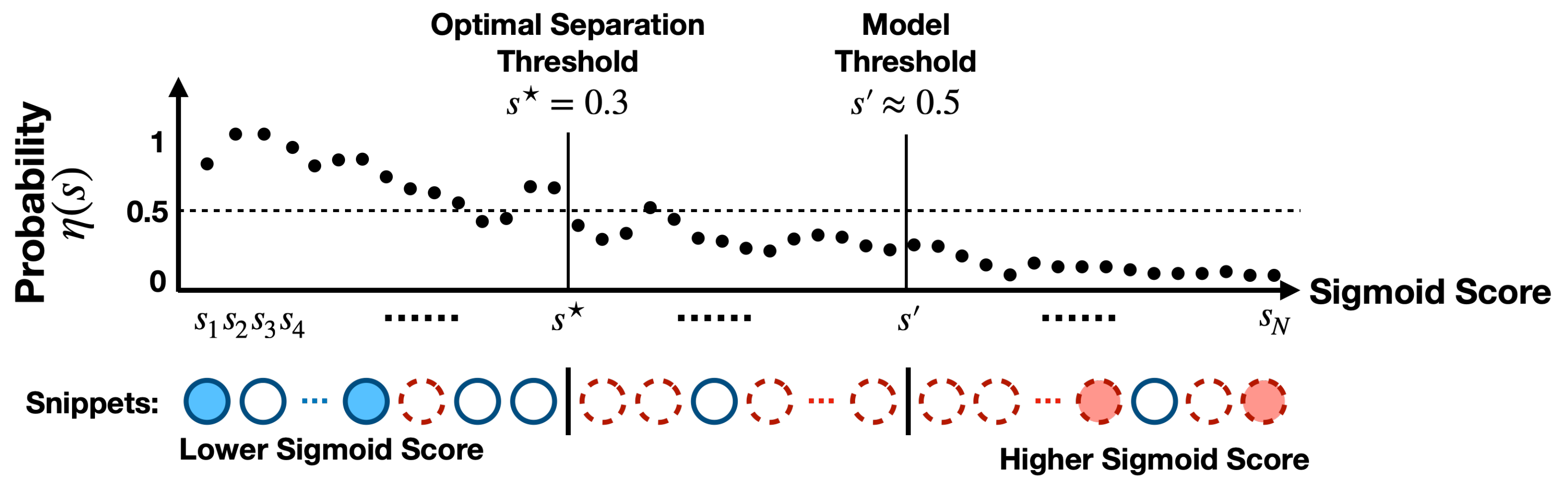}
    \caption{Demonstration of the TRM threshold. Snippets (shown on the bottom) are first ordered based on their predictive sigmoid scores. GPT-4o class labels $0$ and $1$ are represented by the solid or dashed borders. Queried snippets are shaded. Under imbalanced scenarios, sigmoid score of $0.5$ generally will not provide a good indication of where to sample, and will likely result in labeling much more snippets in the majority class. The probability $\mu(s)$ denotes the likelihood of a snippet with sigmoid score $s$ belonging to class 0. The TRM threshold is defined to best separate the two classes of snippets.}
    \label{fig:alg_demo}
    \vspace{-\intextsep}
\end{figure*}

In this paper, we present a novel system SIEVE that addresses these limitations, providing versatile and high-quality data filtering at a fraction of the cost of using high performance, general-purpose models directly. Shown in Figure~\ref{fig:alg_demo}, our method is based on a lightweight classification model in combination with a small number of calls to GPT-4o, effectively reducing the cost per filtering to one call to a lightweight classification model instead of GPT-4o. This is accomplished by training and distilling the lightweight model in the background that learns to mimic the filtering decisions of GPT-4o and eventually handles all/most of the filtering. This background job is optimized via a novel active learning algorithm that further reduces queries to GPT-4o.

Our active learning algorithm operates in a stream-based setting, sequentially and adaptively selecting the most informative text snippets for evaluation by GPT-4o. Inspired by previous work in deep active learning~\citep{zhang2022galaxy,nuggehalli2023direct}, we propose a novel stream-based algorithm designed to tackle scenarios where the data distribution is imbalanced. As all of the filtering decisions are highly imbalanced (see Table~\ref{tab:imbalance}), one of our main contributions is proposing the algorithm to tackle this imbalance issue. Our algorithm is designed to efficiently label a set of more balanced and uncertain examples. As demonstrated in our experiments, SIEVE achieves GPT-4o quality filtering performance at less than 1\% of the cost. In addition, compared to random sampling methods, our active learning algorithm reduces the number of queries to GPT-4o by more than 6x (see Section~\ref{sec:experiments}).
In Section~\ref{sec:analysis}, we provide theoretical analysis of our algorithm, establishing formal proofs of balancedness bounds for active learning in scenarios with imbalanced underlying data distributions. To the best of our knowledge, this represents the first attempt to provide rigorous theoretical guarantees in this context.

The implications of SIEVE are far-reaching, democratizing access to clean, high-quality data for both specialized and general-purpose language models. For specialized domains, SIEVE enables efficient curation of domain-specific datasets from web-scale corpora. For general-purpose LLM training, SIEVE can improve data efficiency by selecting higher quality training examples. By dramatically reducing the cost and complexity of data filtering, SIEVE opens new avenues for researchers and organizations to create both tailored and general-purpose LLMs with improved training data quality. We experimentally validate the effectiveness of SIEVE on both domain-specific filtering tasks and general quality filtering for LLM pretraining data. For domain-specific content filtering, our results suggest SIEVE can achieve the same performance as GPT-4o at less than 1\% of the cost. For quality filtering of general LLM pretraining data, we show that SIEVE can improve upon the existing state-of-art quality filtering mechanisms through the Datacomp-LM challenge~\citep{li2024datacomp}.

\section{A General Purpose Data Filtering System}

In this section we provide a detailed description of the SIEVE system. A visualization of our system can be found in Figure~\ref{fig:sieve_system}.

\textbf{A Bird's-Eye View.} When viewed as a black box, SIEVE processes a web-scale dataset of $N$ text snippets along with a filtering prompt that specifies the criteria for passing or failing each snippet. This prompt specifies the criteria for passing or failing each snippet, similar to how one would instruct any high-performance, general-purpose LLM, such as GPT-4o. From this perspective, SIEVE efficiently categorizes each snippet as `pass' or `fail' based on the provided prompt, with filtering quality comparable to that of GPT-4o.

A straightforward approach would be to apply GPT-4o with the filtering prompt to each text snippet. However, this becomes extremely costly for web-scale datasets containing billions or trillions of tokens. For example, filtering the OpenWebText dataset~\citep{Gokaslan2019OpenWeb} used in this study, which contains 9 billion tokens, would cost approximately \$67,000 using GPT-4o directly. For even larger datasets like the PILE~\citep{gao2020pile}, this cost would increase by at least 1000 times, making it prohibitively expensive.
To overcome this challenge, we utilize an active distillation framework as detailed in the following section.

\subsection{An Active Distillation Framework}
In this section, we describe the inner workings of SIEVE, which employs a lightweight binary classification model trained on GPT-4o's filtering decisions. By leveraging active learning techniques, we selectively gather GPT-4o decisions on a small, informative subset of text snippets from the web-scale dataset. This approach significantly reduces costs while maintaining filtering quality comparable to GPT-4o.

Active learning minimizes annotation costs from expensive sources by selecting the most informative subset of snippets to query for labels. The goal is to train a high-performance model $f$ with minimal annotation cost. In our framework, GPT-4o serves as the expensive annotation source, while we fine-tune a pretrained text encoder model for binary classification on the collected data and annotations to achieve high performance. Active learning strategies collect annotations incrementally, retraining the lightweight classifier model $f$ after labeling every $B$ new snippets. While most deep active learning literature focuses on the pool-based setting \citep{ash2019deep,nuggehalli2023direct,fairstein2024class,lesci2024anchoral}, these algorithms require forward inference of $f$ on the entire dataset at every iteration, incurring high computational costs for large-scale datasets (see Appendix~\ref{ssec:cost} for details). To mitigate this, we apply active learning in the streaming setting, an area well-studied classically but with limited research for neural networks. We specifically designed our algorithm to tackle the imbalance in filtering decisions generated by GPT-4o (see Table~\ref{tab:imbalance}), aiming to query GPT-4o on a more balanced and informative set of text snippets.

Formally, we assume access to a stream of i.i.d. drawn snippets $x_1, x_2, ..., x_N$, all following the same underlying data distribution $\P_X$. We let $S = x_1, ..., x_N$ denote the stream. In practice, we construct the stream by randomly shuffling all snippets in the OpenWebText dataset~\citep{Gokaslan2019OpenWeb}. At time $i$, the active learning algorithm observes $x_i \sim \P_X$ and decides whether to query GPT-4o for its annotation based on the snippet's informativeness to model $f$. If queried, we obtain the corresponding filtering decision from GPT-4o, which we denote by $y_{GPT}(x_i) \in \{0, 1\}$. Here, the randomness comes from the nondeterministic nature of GPT-4o's response. After every $B$ new annotations, we fine-tune the lightweight text encoder model from its pretrained checkpoint to obtain an updated model $f$. The distillation process terminates after a total annotation budget of $T$ snippets. The final model $f$ fine-tuned on queried snippets is then applied to filter the entire web-scale dataset.

\section{Active Learning Algorithm}

In this section, we present a novel stream-based active learning algorithm for class-imbalanced scenarios. We use a modified version of the uncertainty sampling strategy~\citep{lewis1994sequential,tong2001support, balcan2006agnostic,settles2009active, kremer2014active}. In our setting, uncertainty sampling labels snippets $x_i$ that have predictive sigmoid score around $0.5$, i.e. $f(x_i) \approx 0.5$, as these  are believed to be the most informative data to be labeled. However, our binary classification tasks of data filtering is naturally imbalanced as shown in Table~\ref{tab:imbalance}. Previous active learning work by \citet{zhang2022galaxy} and \citet{nuggehalli2023direct} observed that the threshold of $0.5$ is generally biased towards the majority class under imbalanced scenarios, which means that most of the snippets selected by uncertainty sampling will be in the majority class.  This translates into poor performance in detecting snippets from the target minority class.  To combat this, our approach instead aims to find a threshold near where the majority and minority classes are equiprobable, which we will refer to as an \emph{True Risk Minimizer (TRM) threshold}.  Selecting and labeling snippets around the TRM threshold yields a labeled dataset that is more balanced and includes uncertain snippets. In a nutshell, our algorithm (detailed in Algorithm~\ref{alg:active}) alternates between
\begin{enumerate}
    \item Spending $B$ budget in finding and labeling close to the TRM threshold;

    \item Fine-tuning a new text encoder model $f$ on all labeled snippets thus far.
\end{enumerate}
More concretely as shown in Figure~\ref{fig:alg_demo}, when ordering snippets according to their sigmoid scores, labeling snippets around the threshold of $0.5$ can often result in selecting and labeling of most of snippets from a single (majority) class. Our proposed alternative, the TRM threshold, minimizes the expected number of class $1$ snippets to its left and class $0$ snippets to its right. Thus, the TRM threshold minimizes the expected number of misclassifications. 
Formally, let $\eta_{j}$ denote the probability of snippet $x_j$ belonging to class $0$
$$\eta_j \ := \ \P\big(y_{\mbox{\tiny GPT}}(x_{j})= 0\big)$$ 
where the possible randomness is with respect to GPT-4o's response as well as the distribution underlying the snippet $x_j \sim P_X$. The distribution $P_X$ is unknown, so $\{\eta_{j}\}$ are unknown as well. 
For model $f$ and a stream of i.i.d. text snippets $x_1, ..., x_N \sim \P_X$, the \emph{True Risk Minimizer (TRM) threshold} is then the sigmoid score $s^\star \in \{0, f(x_1),\dots,f(x_N)\}$, where
% \begin{align}
%     r^\star \ := \ \argmin_{0 \leq k \leq N} \left(\sum_{j=1}^k (1 - \eta_{(j)}) + \sum_{j=k+1}^N \eta_{(j)}\right).
% \end{align}
\begin{equation}
    s^\star \ := \ \argmin_{\substack{s \in \\\{0, f(x_1),\dots,f(x_N)\}}} \bigg(\sum_{j: f(x_j)\leq s} (1 - \eta_j) + \sum_{j: f(x_j)>s} \eta_j\bigg).
\end{equation}
The optimization objective above can be viewed as the expected risk (probability of error) of a finite set of threshold classifiers with threshold locations at $\{0, f(x_1),\dots,f(x_N)\}$.
A threshold classifier at $s$ categorizes snippets into class $0$ if their sigmoid score is less than or equal to $s$, and into class $1$ if their score is greater than $s$. The TRM threshold can also be viewed as where the two classes best separate from each other. Snippets around this threshold are therefore truly uncertain to the neural network model. Moreover, in Section~\ref{sec:analysis}, we will provide a novel theoretical analysis demonstrating labeling around the TRM threshold also improves the balancedness of the queried snippets, alleviating the data imbalance issues when training the lightweight model $f$.  Of course, the TRM threshold is unknown (because the probabilities $\{\eta_{j}\}$ are unknown), so we employ an efficient active learning procedure to identify a threshold close to it.

\setlength{\textfloatsep}{.5\intextsep}
\begin{algorithm}[ht!]
    \begin{algorithmic}
    \STATE \textbf{Input: } Data Stream $S$, labeling function $y_{GPT}$ based on GPT-4o and specified filtering prompt, batch size $B$, total budget $T$ and confidence level $\delta$.
    \STATE \textbf{Initialize: } Query $x_1$, ..., $x_B \overset{iid}{\sim} \P_X$ to form the initial labeled set $L = \{x_i, y_i\}_{i=1}^B$.

    \FOR{$r = 1, ..., \frac{T}{B}$}
        \STATE Fine-tune pretrained encoder model (e.g., T5) on the latest labeled set $L$ to form $f: S \rightarrow [0, 1]$.
        \STATE Initialize confidence set $\underline{\mu}, \bar{\mu} \leftarrow 0, 1$, counter $t \leftarrow 0$.
        \STATE Let $j$ index the head of the stream $S$, $x_j$.
        
        \WHILE{$|L| < (s + 1) B$}

        \STATE \underline{\textbf{Find optimal separation threshold for $f$.}}
        \STATE Receive next snippet in the stream $S$, $x_{j+t} \sim \P_X$.

        \IF{$f(x_{j+t}) \in [\underline{\mu}, \bar{\mu}]$}
            \STATE Query GPT-4o for label $y_{j+t}$ by $y_{GPT}$, and insert to set $L \leftarrow L \cup \{x_{j+t}, y_{j+t}\}$.
        \ENDIF

        \IF[\underline{\textbf{Update confidence interval.}}]{$t \in 2^{\N^+}$}
            \STATE Store previously computed sigmoid scores $F_t \leftarrow \{0, f(x_j), ..., f(x_{j + t})\}$.
        
            \STATE Compute the empirical TRM threshold as $\widehat{s}_t \leftarrow \min_{s \in F_t} \widehat{\mc{L}}_{t}(s)$, where \\ 
            $\widehat{\mc{L}}_{t}(s) := \frac{1}{t+1}(\sum_{i\in[j, j+t]: f(x_{i}) \leq s} \1\{y_{i} \neq 0\} + \sum_{i\in[j, j+t]: f(x_{i}) > s} \1\{y_{i} \neq 1\})$.

            \STATE Update $\underline{\mu}, \bar{\mu}$ be the smallest and largest thresholds $s\in F_t$ such that
            \begin{align*}
                &\widehat{\mc{L}}_{t}(s) - \widehat{\mc{L}}_{t}(\widehat{s}_t) \leq \beta_{t+1}^2 / 2 + \\
                &\beta_{t+1}\cdot\sqrt{(|\{s'\in F_t: \underline{s} \leq s' \leq \bar{s}\}|-1) / t}
            \end{align*}
            where $\underline{s} = \min(s, \widehat{s}_t)$, $\bar{s} = \max(s, \widehat{s}_t)$ and $\beta_{t+1} = \sqrt{2\log(2\log_2(t+1)^2N^2/\delta)/(t+1))}$ are chosen so that the TRM threshold lies in the updated confidence threshold with probability of at least $1 - \delta / \log_2(N)$.
        \ENDIF

        \STATE Update stream counter $t \leftarrow t+1$ over snippets $S$. 
        % \COMMENT{\textbf{\underline{Loop over snippets in S.}}}
        \ENDWHILE

    \ENDFOR

    \STATE \textbf{Return:} Text classifier model $f$ finetuned on $L$. $f$ is then used for filtering on the entire dataset. 
    \end{algorithmic}
    \caption{Stream-Based Class-Balancing Active Learning}
    \label{alg:active}
\end{algorithm}

Our algorithm, shown in Algorithm~\ref{alg:active}, applies agnostic active learning techniques~\citep{dasgupta2007general,jamieson2022interactive,katz2021improved} to threshold classifiers. It focuses on identifying the TRM threshold. The algorithm initializes by querying the first $B$ snippets from the stream to create a labeled set $L$. Each iteration begins with fine-tuning classifier $f$ on $L$. To estimate the TRM threshold $s^\star$, we maintain a high-confidence interval $[\underline{\mu}, \bar{\mu}]$ while annotating stream snippets. Snippets with predictive sigmoid scores within this interval are queried using GPT-4o. We update the confidence interval on a geometric schedule, every $2^i$ snippets, using empirical estimates $\widehat{s}_t$ of the optimal threshold. This interval, centered around the empirical estimate, ensures $s^\star \in [\underline{\mu}, \bar{\mu}]$ with at least $1 - \delta$ probability across all $t=2^i$ for the current iteration. For constructing the confidence interval, we employ an empirical Bernstein bound, as introduced in~\citet{jamieson2022interactive}. This approach differs from the original CAL algorithm~\citep{dasgupta2007general}, which uses a uniform convergence bound requiring synchronous querying. Our method allows for parallel GPT-4o queries, significantly reducing computational time in practice.

\section{Theoretical Analysis} \label{sec:analysis}
In this section, we first provide formal guarantee of the performance of our algorithm. We then proceed to analyze the balancedness of the snippets our algorithm queries, showing its improvement in collecting a more balanced set of snippets. Note that all of our analysis are conducted for any single iteration $s$ in Algorithm~\ref{alg:active}.
%\jifan{All of the analysis below applies for any single iteration $r$ in Algorithm~\ref{alg:active}. For ease of notation, we use $x_1, \dots, x_t\overset{iid}{\sim} \P_X$ to denote the examples $x_{j}, ..., x_{j+t-1} \overset{iid}{\sim} \P_X$ in each iteration of Algorithm~\ref{alg:active}.}
%Denote the empirical risk function over these snippets as 
%$$\widehat{R}_t(s) \ = \ \frac{1}{t}\left(\sum_{i\leq t \, : \,  f(x_i) \leq s} \1\{y_i \neq 0\} + \sum_{i\leq t \, :\,  f(x_i) > s} \1\{y_i\neq 1\}\right). $$
%Since the active learning algorithm labels every snippet in the region of disagreement, the minimizer $\widehat s_t = \arg\min_s \widehat R_t(s)$ is computable.
Define the true risk function over all snippets $1,\dots,N$ at threshold $s$ as
$$R(s) \ = \  \frac{1}{N}\left(\sum_{i\leq N: f(x_i) \leq s} (1 - \eta_{i}) + \sum_{i\leq N: f(x_i) > s} \eta_{i}\right) \ .$$
Recall the goal is to find a threshold near $s^* = \arg\min_s R(s)$.  This learning problem over the discrete set of threshold classifiers at thresholds $f(x_1)$, ..., $f(x_N)$ can be exactly represented in the framework of \citet{jamieson2022interactive}, leading to Algorithm~\ref{alg:active} and the following bound.

%For any value $1\leq t \leq N$, let $\pi_t$ denote the permutation of $\{1,\dots,t\}$ such that $f(x_{\pi_t(1)}) \leq \cdots \leq f(x_{\pi_t(t)})$.
% \rob{I think it's important that we assume the snippets are iid, and so we should state this again here.}
% \begin{theorem}[Empirical Risk Bound]
% For each $t$ and any thresholds $s_1 \leq s_2 \in \{f(x_1),\dots,f(x_t)\}$, with probability at least $1-\delta / \log_2(N) / 2$
% \begin{align*}
%     \widehat{R}_t(s_1) - \widehat{R}_t (s_2) \ \leq \ R(s_1) - R(s_2) \ + \ \beta_t\sqrt{\Delta(s_1,s_2) / t} \ + \ \beta_t^2 / 2.
% \end{align*}
% where $\Delta(s_1,s_2) = |\{i \, : \, s_1 \leq f(x_i) \leq s_2\}|-1$, the number of thresholds in $\{f(x_1),\dots,f(x_t)\}$ falling between $s_1$ and $s_2$.
% \end{theorem}
% \begin{proof}
% The empirical and true risks are defined for thresholds in the set of sigmoid scores $\{f(x_1),\dots,f(x_N)\}$.
% The theorem statement bounds differences in empirical risks in terms of the differences of true risks at threshold sigmoid scores at $s_1$ and $s_2$. The bound is a simple application of the empirical Bernstein inequality \cite{maurer2009empirical}.
% \end{proof}

\begin{theorem}[\citet{jamieson2022interactive}]
    During iteration $r$ of Algorithm~\ref{alg:active}, given the classifier model $f$, with probability at least $1 - \delta$, both $R(\underline{\mu}) - R(s^\star)$ and $R(\bar{\mu}) - R(s^\star)$ are upper bounded by
    \begin{align}
        c_0 R(s^\star) + c_1\beta_t \sqrt{R(s^\star)} + c_2\beta_t^2.
    \end{align}
    for all the confidences intervals $[\underline{\mu}, \bar{\mu}]$ updated at time $t \in 2^{N^+}$. Here, $c_0, c_1$ and $c_2$ are some universal constants.
\end{theorem}
The theorem suggests that the gap in the true risk of the confidence intervals shrinks roughly on the scale of $\frac{1}{\sqrt{t}}$ over time since $\beta_t = \widetilde{O}(\frac{1}{\sqrt{t}})$. For large $t$, this gap goes to $0$.

\begin{definition}[Score Re-Ordering]
Let $\pi$ denote the permutation of $\{1,\dots,N\}$ such that $f(x_{\pi(1)}) \leq \cdots \leq f(x_{\pi(N)})$.
\end{definition}

% \rob{seems easier not to reorder.  ultimately we are considering all the $\eta(X)$ values corresponding to $X$ with scores $f(X)$ close to $s^*=f(X^*)$. Then we want to argue that if $|f(X)-f(X^*)|$ is small so is the difference $|\eta(X)-\eta^*(X)|$.  So aren't we really just saying that $|\eta(X)-\eta^*(X)|\leq L |f(X)-f(X^*)|$?}
\begin{definition}[Discrete Smoothness]
    Let $L = \max_{j\in [N-1]} |\eta_{\pi(j)} - \eta_{\pi(j+1)}|$ denote the maximum change in probabilities $\eta_{(j)}$. This mirrors the Lipschitz smoothness in a discrete fashion. Note we always have $L\leq 1$.
\end{definition}

Without loss of generality, assume class $0$ to be the minority class. The expected imbalance ratio is then $\frac{\sum_{j=1}^N \eta_{j}}{\sum_{j=1}^N 1-\eta_{j}} < 1$. If $s^\star = 0$, it means even the lowest sigmoid score snippets are likely to be in the majority class. In this case, a reasonable strategy is to simply query the lowest sigmoid score snippets, which is exactly what our algorithm does.

\begin{theorem}[Balancedness of Labeled Snippets]
    Assume class 0 is the minority class and $s^\star \neq 0$. Consider an interval of scores $[\underline{\mu},\bar{\mu}]$ with $s^\star \in [\underline{\mu},\bar{\mu}]$. Let the corresponding gaps in risk denoted by $\gamma_0 := R(\underline{\mu}) - R(s^\star) > 0$ and $\gamma_1 := R(\bar{\mu}) - R(s^\star) > 0$. When labeling snippets indexed within this interval uniformly at random, the imbalance ratio $\lambda(\underline{\mu},\bar{\mu})$ between the minority class and the majority class must satisfy
    \begin{align*}
        & \lambda(\underline{\mu},\bar{\mu}) = \frac{\sum_{j: f(x_j)\in[\underline{\mu},\bar{\mu}]}\eta_{j}}{\sum_{j: f(x_j)\in[\underline{\mu},\bar{\mu}]}1-\eta_{j}} \\
        & \geq 1 - \min(\frac{N\bar{\gamma} + LN}{1.5 - 2L}, \sqrt{L}\cdot\frac{N\bar{\gamma} + LN + 1}{(1-L)\sqrt{N\underline{\gamma}}})
    \end{align*}
    where $\underline{\gamma}:=\min(\gamma_0, \gamma_1)$ and $\bar{\gamma}:=\max(\gamma_0, \gamma_1)$, with $L, \underline{\gamma}, \bar{\gamma}< 1$. This implies, 
    
    (a) if $\bar{\gamma} \rightarrow 0$ and $L \rightarrow 0$, $\lambda(\underline{\mu},\bar{\mu}) \geq 1 - \frac{N\bar{\gamma} + LN}{1.5 - 2L} \rightarrow 1$ (perfect balance);

    (b) if $\underline{\gamma} \geq \frac{c}{N}$ for some constants $c > 0$, then $\lambda(\underline{\mu},\bar{\mu}) \geq 1 - \sqrt{L}\cdot\frac{N\bar{\gamma}+LN + 1}{c(1-L)}$. When $L \rightarrow 0$, we again recovers $\lambda(\underline{\mu},\bar{\mu}) \rightarrow 1$ (perfect balance).
\end{theorem}

\textit{Proof sketch.}
Let $A = \sum_{j: f(x_j)\in[\underline{\mu},\bar{\mu}]}\eta_{j}$ and $B=\sum_{j: f(x_j)\in[\underline{\mu},\bar{\mu}]}1 - 2\eta_{j}$, we can rewrite the imbalance ratio into $\lambda(\underline{\mu},\bar{\mu}) = \frac{A}{A + B}$. Since $N(\gamma_1 - \gamma_0) = \sum_{j:f(x_j)\in (\underline{\mu}, \bar{\mu}]} 1 - 2\eta_j \geq B - LN$, we can lower bound the balancedness by $\frac{A}{A + N(\gamma_1 - \gamma_0) + LN}$. We then need to prove that $A$ is sufficiently large relative to $N(\gamma_1 - \gamma_0) + LN$.

To prove this, we note that the $\eta$ values around the optimal separation threshold $s^\star$ are close to $.5$. Therefore, by the smoothness condition, when $L$ is small, $\eta_j \approx .5$ for all $f(x_j) \in [\underline{\mu},\bar{\mu}]$, so $A$ roughly scales linearly in the number of elements within $[\underline{\mu},\bar{\mu}]$. We can prove that there are at least $O(\sqrt{\frac{N\bar{\gamma}}{L}})$ elements in this confidence interval. $A$ therefore follows a similar scale, which is much greater than $N(\gamma_1 - \gamma_0) + LN$ when $L\rightarrow 0$. Under such case, we recover a balancedness bound of $1$, i.e. the annotated snippets are perfectly balanced. See Appendix~\ref{apx:proof} for complete proof.

\section{Experiments} \label{sec:experiments}
\begin{table*}[t]
    \centering
    \scalebox{0.8}{
    \begin{tabular}{llccccc}
    \toprule
    Filter & Method & \shortstack{Bal. Accuracy\\(GPT-4o as GT)} & \shortstack{Human Preference\\Over GPT-4o} & \shortstack{\#Queries to \\ GPT-4o} & \shortstack{Lightweight \\Model Cost} & Total Cost\\
    \midrule
    \multirow{3}{*}{Climate} & GPT-3.5-Turbo & 88.9\% & --- & 13.5M & \$0 & \$13,000\\
    & GPT-4o & \textbf{96.6\%}$^\ast$ & --- & 13.5M & \$0 & \$67,000\\
    & SIEVE (Ours) & \textbf{96.7\%} & --- & 7.5K & $\$80$ & $\$120$\\
    \midrule
    \multirow{3}{*}{AI} & GPT-3.5-Turbo & 58.22\% & --- & 13.5M & \$0 & \$13,000\\
    & GPT-4o & \textbf{95.5\%}$^\ast$ & --- & 13.5M & \$0 & \$67,000\\
    & SIEVE (Ours) & \textbf{95.7\%} & --- & 6K & $\$80$ & $\$110$ \\
    \midrule
    \multirow{3}{*}{Politics} & GPT-3.5-Turbo & 84.13\% & --- & 13.5M & \$0 & \$13,000\\
    & GPT-4o & \textbf{95.6\%}$^\ast$ & --- & 13.5M & \$0 & \$67,000\\
    & SIEVE (Ours) & \textbf{95.6\%} & --- & 60K & $\$170$ & $\$470$\\
    \midrule
    \multirow{3}{*}{\shortstack[l]{Mainstream}} & GPT-3.5-Turbo & 76.11\% & --- & 13.5M & \$0 & \$13,000\\
     & GPT-4o & 92.4\%$^\ast$ & $50\%$ & 13.5M & \$0 & \$67,000\\
    & SIEVE (Ours) & 91.0\% & $\textbf{54\%}$ & 100K & $\$300$ & $\$800$ \\
    \midrule
    \multirow{3}{*}{Quality} & GPT-3.5-Turbo & 74.5\% & --- & 13.5M & \$0 & \$13,000\\
    & GPT-4o & 88.2\%$^\ast$ & 50\% &  13.5M & \$0 & \$67,000\\
    & SIEVE (Ours) & 86.3\% & \textbf{53\%} & 60K & $\$170$ & $\$470$\\
    \bottomrule
    \end{tabular}
    }
    \caption{Performance results of applying SIEVE on five highly specialized filters~(see Appendix~\ref{apx:prompts}). SIEVE can match or exceed GPT-4o's quality in terms of balanced accuracy and human preference. On the other hand, SIEVE saves more than 99\% of the cost compared to using GPT-4o. The lightweight model cost includes active learning, model training and inference on OpenWebText. See Section~\ref{sec:experiments} for experiment details and Appendix~\ref{ssec:cost} for cost breakdown. $^\ast$We assess GPT-4o's accuracy by measuring output consistency between identical API calls using greedy decoding (temperature = 0). Some inconsistency persists, possibly due to hardware non-determinism. This may be amplified when using our CoT prompts.}
    \label{tab:result}
    \vspace{-\intextsep}
\end{table*}

In this section, we present our experiments conducted for the five highly customized data filters in Table~\ref{tab:result}: politics, climate, AI, mainstream knowledge and text quality filters. These filters are applied to the OpenWebText dataset~\citep{Gokaslan2019OpenWeb}, which is divided into around 13.5M snippets of $1024$ tokens. The first three filters identify text snippets related to a particular topic, with highly detailed specifications of filtering prompts to GPT-4o about the subdomain of topics that should be included. The mainstream knowledge prompt aims to exclude any obscure and niche content determined by GPT-4o. Lastly, the quality filter aims to identify text snippets that are considered high quality by GPT-4o. See Appendix~\ref{apx:prompts} for prompts.

Throughout our experiments, we use the encoder part of pretrained T5-large~\citep{raffel2020exploring} as the lightweight model. We also conduct an ablation study that uses the DeBERTa-v2-xlarge model~\citep{he2020deberta}, and found no significant differences in outcomes (see Appendix~\ref{apx:model_ablation}, Table~\ref{tab:model_ablation}). A linear layer is attached to the encoder model for binary classification. The classification model has less than $770$M parameters, orders of magnitude smaller than GPT-4o. To mitigate the imbalanced nature of the snippets shown in Table~\ref{tab:imbalance}, we utilize the focal loss~\citep{lin2017focal}. We refer the readers to Appendix~\ref{apx:training} for more training details.

\subsection{Results}

\textbf{GPT-4o-Based Evaluation.} Table~\ref{tab:result} demonstrates that, with significantly lower total cost, SIEVE achieves comparable balanced accuracy to GPT-4o on climate, AI and climate filters, while closely matching its performance on mainstream knowledge and text quality filters. We evaluated these filters using a set of test snippets randomly sampled from OpenWebText. Ground truth labels were established through individual GPT-4o API calls for each snippet.
To assess GPT-4o's performance against this ground truth, we conducted additional API calls using identical prompts for each test snippet, effectively measuring the model's self-consistency rate. We observed inherent noise in GPT-4o's decisions, even when using greedy decoding (temperature set to $0$). This variability likely stems from non-deterministic factors in the hardware infrastructure.
While we employed chain-of-thought (CoT) reasoning in our filtering prompts (detailed in Appendix~\ref{apx:prompts}) to enhance decision quality, the compounding noise from each generated token may have contributed to the noticeable inconsistencies.

\textbf{Human Evaluation.} In the above, we compared our distilled lightweight model's accuracy to GPT-4o for both mainstream and quality filters. When evaluated by GPT-4o itself, our model's performance appeared lower, raising the question: Is our lightweight model actually worse, or is GPT-4o biased when judging its own decisions? To investigate, we conducted a human evaluation. For each filter, we randomly selected 100 text snippets where GPT-4o and our lightweight model disagreed. We then recruited two groups of 13 annotators per filter to manually assess these challenging cases. The ground truth for each snippet was determined by the majority vote of the annotators, allowing us to compare both models' performance against human judgment.
As shown in the fourth column of Table~\ref{tab:result}, we see even a slight edge of our lightweight model over GPT-4o when judged by human. This suggests our lightweight model is at least comparable to the decisions made by GPT-4o, while incurring much less computational cost when filtering web-scale datasets.

\textbf{Cost Breakdown and Comparison.}
Our lightweight model cost encompass active learning, model training and inference cost across the entire SIEVE pipeline. Please see Appendix~\ref{ssec:cost} and its Table~\ref{tab:cost_comparison} for more details and comparisons.
To summarize our findings, active distillation in SIEVE offers a more cost-effective approach compared to random distillation in achieving equivalent model performance. This cost advantage becomes even more pronounced when considering more expensive teacher models, such as the recently released o1, which significantly increases query costs and potentially dominates the total expenses. Given that active distillation substantially reduces the number of required queries, the cost disparity between active and random distillation methods is expected to widen further, especially when utilizing more advanced and costly teacher models. We further provide a discussion of our choice of stream-based vs pool-based active learning algorithm in Appendix~\ref{ssec:cost}.

\subsection{Active vs Random Distillation}
\begin{figure*}
\begin{subfigure}[t]{.33\linewidth}
    \includegraphics[trim={0 1.2cm 0 1cm}, width=\linewidth]{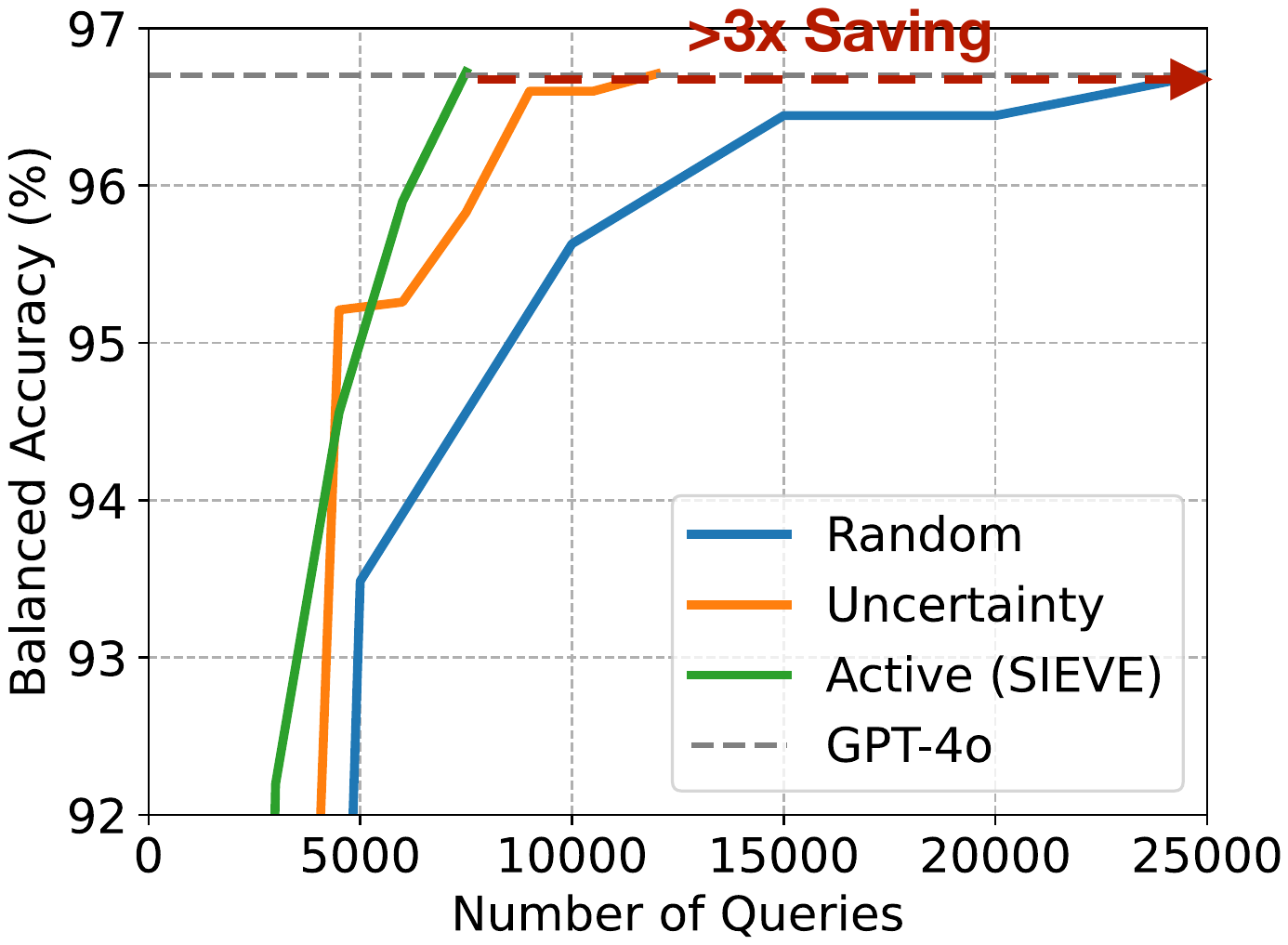}
    \caption{Climate filter accuracy}
    \label{fig:climate_comparison}
\end{subfigure}
\begin{subfigure}[t]{.33\linewidth}
    \includegraphics[trim={0 1.2cm 0 1cm}, width=\linewidth]{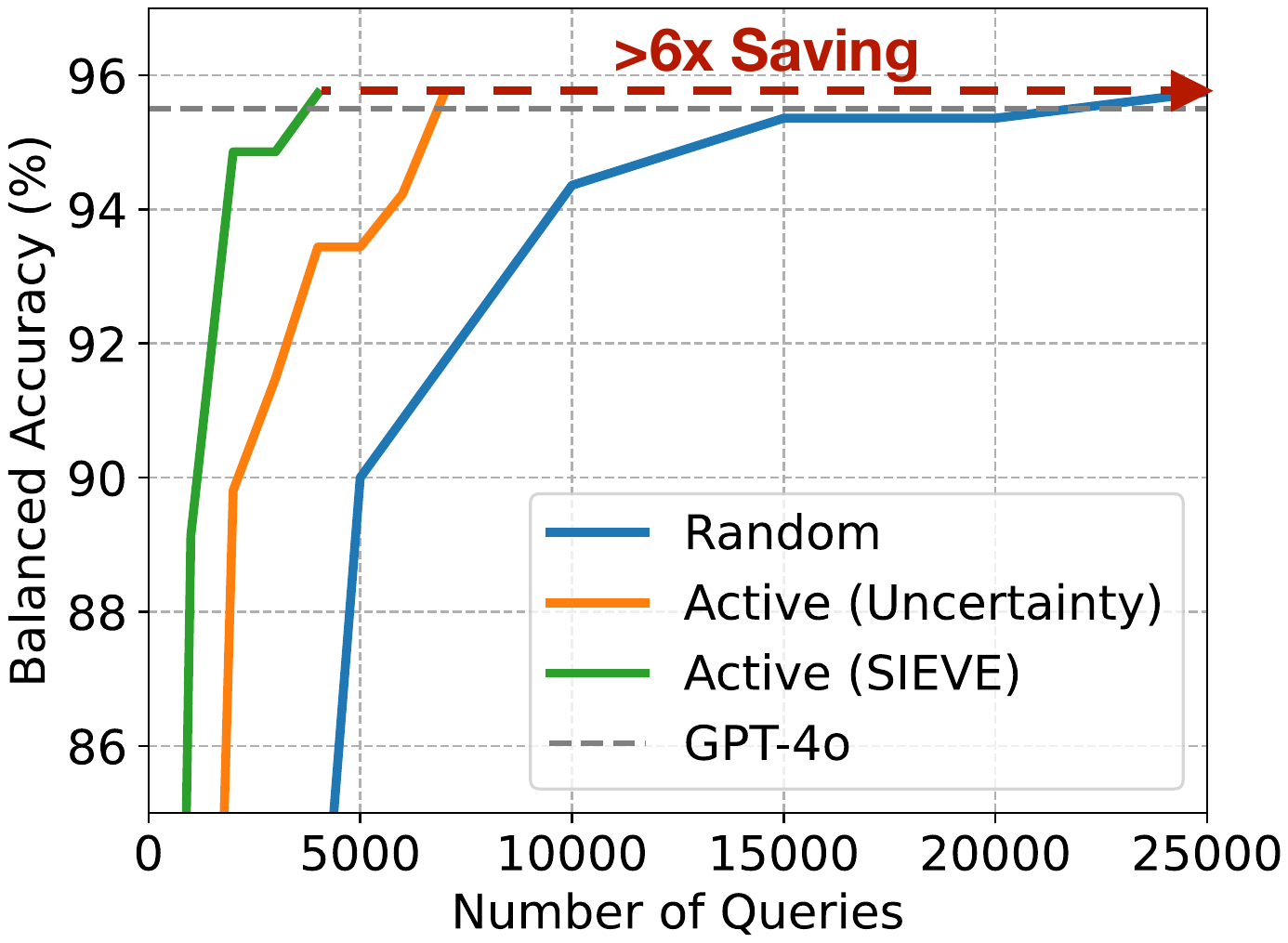}
    \caption{AI filter accuracy}
    \label{fig:ai_comparison}
\end{subfigure}
\begin{subfigure}[t]{.33\linewidth}
    \includegraphics[trim={0 1.2cm 0 1cm}, width=\linewidth]{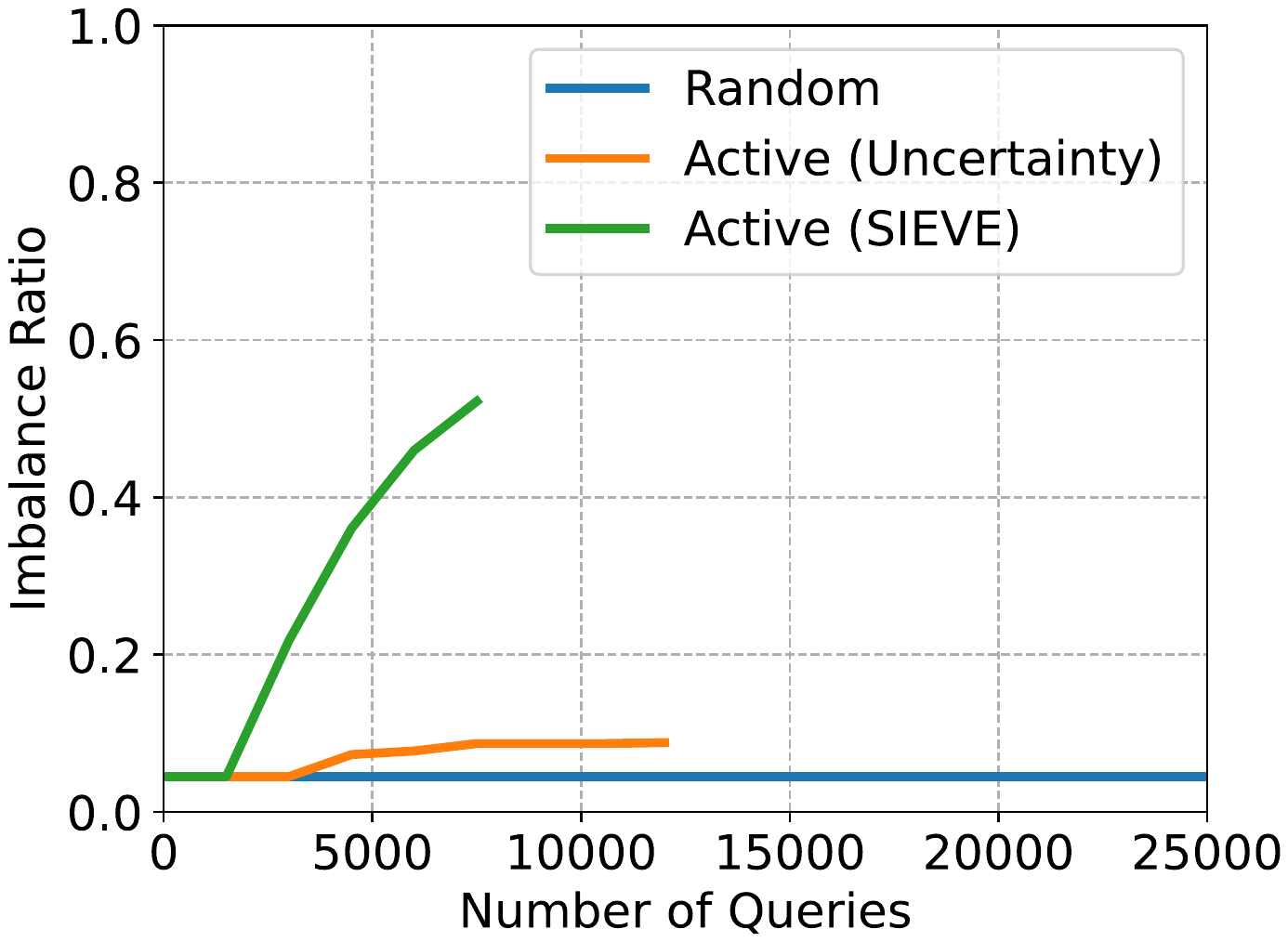}
    \caption{Imbalance ratio of queries, climate filter}
    \label{fig:climate_imb}
\end{subfigure}
\vspace{-.5\intextsep}
\caption{\textbf{Active vs Random Distillation}: Performance of the distilled lightweight model across different number of queries made to GPT-4o. With active learning algorithm proposed in SIEVE, we can save the number of queries to GPT-4o by more than 5x for the politics filter and more than 3x for the climate filter. We also observe significant query saving against the classic uncertainty sampling algorithm.}
\label{fig:active_random}
\vspace{-\intextsep}
\end{figure*}

In this section, we compare the effectiveness of active learning algorithms in Algorithm~\ref{alg:active} and the classic uncertainty sampling algorithm~\citep{lewis1994sequential,tong2001support, settles2009active, balcan2006agnostic, kremer2014active} against random querying for distilling our lightweight model. 
% The random distillation strategy involves querying GPT-4 on a predetermined number of randomly selected snippets from the OpenWebText dataset, followed by fine-tuning the encoder model on the queried data. 
As illustrated in Figure \ref{fig:active_random}, our comparison of the lightweight model's performance reveals significant advantages for active distillation from SIEVE over random sampling. Specifically, SIEVE demonstrates remarkable efficiency, requiring more than three and six times fewer GPT-4o queries compared to random distillation. Figure~\ref{fig:climate_imb} also demonstrates our active learning algorithm's effectiveness in querying a much more balanced set of snippets. Moreover, we also consistently see a 2-3x improvement from SIEVE when comparing against uncertainty sampling.
As inference cost of state-of-art LLMs increases, active distillation could play an increasingly important role in SIEVE.

\subsection{Datacomp-LM with SIEVE's Quality Filter}
\begin{table}[t]
    \centering
    \scalebox{.9}{
    \begin{tabular}{lccc}
        \toprule
         Method & Core& Extended & MMLU \\
         \midrule
         DCLM-baseline (reprod.) & .1761 & .0946 & .2389\\
         Fineweb-edu (reprod.) & .1685 & .0846 & .2332 \\
         DCLM-baseline + SIEVE & \textbf{.1790}& \textbf{.1008} & \textbf{.2599}\\
         \bottomrule
    \end{tabular}}
    \caption{Performance of different pretraining datasets on the Datacomp-LM filtering challenge on 400M-1x scale. A fixed budget of 8.2B training tokens and model training procedure is given by the challenge. We observe quality filters trained by SIEVE with GPT-4o can further boost the pretraining data selection performance.}
    \label{tab:dclm}
\end{table}

We evaluate \textsc{SIEVE} on the Datacomp-LM benchmark~\citep{li2024datacomp} in the 400M-1x scale challenge, which requires filtering a 240T token Common Crawl pool to 8.2 billion training tokens. A language model with 400M parameters and fixed architecture is trained on this filtered dataset and evaluated using multiple metrics from~\citet{li2024datacomp}. Currently, the Datacomp-LM's \emph{DCLM baseline} and \citet{lozhkov2024fineweb-edu}'s \emph{Fineweb-edu} lead the challenge. The DCLM baseline employs a fastText-based quality filtering model~\citep{joulin2016bag}, which was trained to identify high-quality training data similar to those in OpenHermes 2.5~\citep{teknium2023openhermes} and r/ExplainLikeImFive subreddits. Fineweb-edu~\citep{lozhkov2024fineweb-edu} trains a quality filtering model based on Llama3's annotated data.

In our experiment, we train a quality filter distilled from GPT-4 using SIEVE and apply it to further filter the DCLM baseline subset. As shown in Table~\ref{tab:dclm}, our approach outperforms both the DCLM baseline and Fineweb-edu. We defer direct filtering from DCLM-pool and DCLM-RefineWeb for future work due to computational constraints. Our reproduced results of DCLM-pool and Fineweb-edu slightly underperform the Datacomp-LM leaderboard scores across three random trials of selecting 8.2B token subsets from the 4T token datacomp-baseline set. We attribute this discrepancy to potential overfitting in the original authors' data selection process through repeated model retraining and selecting the results of the best subset.

\vspace{-1mm}
\section{Related Work}
\textbf{Data Filtering for Large Language Models.} Data curation is fundamental to the development of LLMs \citep{longpre2023pretrainer, zhou2023oasis}. As interest in domain-specific LLMs grows, the need for extensive, relevant data collection becomes increasingly important. For a comprehensive overview of existing datasets, we direct readers to \citet{raffel2020exploring,gao2020pile,liu2024datasets}.
Current methods for acquiring domain-specific data predominantly rely on a few established large-scale databases, including textbooks \citep{gunasekar2023textbooks}, code repositories \citep{muennighoff2023octopack,gao2020pile}, medical literature, and other specialized sources \citep{gao2020pile,cheng2023adapting}. However, for rapidly evolving topics like the 2024 presidential election, climate change and artificial intelligence, relevant information is often dispersed across the internet, making comprehensive data collection challenging.
Our approach involves training lightweight, task-specific data filtering models distilled from GPT-4. These models are then applied to web-scale datasets to identify pertinent information across various domains.
Existing data filtering techniques span a wide range, from basic rule-based methods utilizing sentence-level statistical features \citep{rae2021scaling,yang2019xlnet,laurenccon2022bigscience,zhang2022opt} to sophisticated filters leveraging pretrained neural networks for text quality~ \citep{brown2020language,du2022glam,chowdhery2023palm,touvron2023llama, enomoto2024investigating,qian2024understanding,li2024datacomp} and toxicity~\citep{lees2022new,friedl2023dis} filtering.
To our knowledge, this work represents the first attempt to develop domain-specific data filtering models adaptable to a diverse array of filtering requirements and specialized domains.

\textbf{Active Learning.}
Active learning is a strategy aimed at reducing data annotation costs by selectively choosing which examples to label. Traditional approaches iteratively update machine learning models based on newly labeled data, using various informativeness metrics to guide the selection process. These metrics typically include uncertainty~\citep{lewis1994sequential,tong2001support, settles2009active, balcan2006agnostic, kremer2014active,gal2017deep,ducoffe2018adversarial,beluch2018power}, diversity~\citep{sener2017active,geifman2017deep,citovsky2021batch}, and expected model change~\citep{ash2019deep,ash2021gone,wang2021deep, elenter2022lagrangian,mohamadi2022making}. However, most existing methods are designed for pool-based settings with balanced data distributions, which may not be suitable for all real-world scenarios.

In contrast, our work focuses on stream-based algorithms capable of handling data imbalance, an area that has received limited attention in deep learning contexts. While pool-based algorithms have shown success in addressing class imbalance~\citep{aggarwal2020active, kothawade2021similar, emam2021active, zhang2022galaxy, coleman2022similarity, jin2022deep, cai2022active,nuggehalli2023direct,zhang2024algorithm,lesci2024anchoral,fairstein2024class}, stream-based approaches for deep neural networks remain understudied. The recent work by \citet{saran2023streaming} introduces an online volume sampling technique for annotating diverse sets of examples in the representation space. Representation diversity, however, has been shown to struggle on class-imbalanced data distribution under the pool based setting~\citep{zhang2022galaxy,nuggehalli2023direct,lesci2024anchoral,fairstein2024class}. Specifically these studies suggest that sampling diversely in the representation space does not necessarily improve the class-balancedness of the annotated examples. To address this gap, we propose what we believe to be the first stream-based algorithm specifically designed for class imbalance scenarios in active learning.

Lastly, we provide an extended literature review of agnostic active learning and knowledge distillation in Appendix~\ref{apx:lit}.

\vspace{-1mm}
\section{Conclusion, Limitations and Future Work}
In this paper, we introduced SIEVE, demonstrating the feasibility of achieving GPT-4o quality data filtering across a diverse range of user-specified filtering prompts. Our comprehensive study showcases the effectiveness of SIEVE in curating large-scale, high-quality datasets for language model training at a fraction of the cost of existing techniques. The experimental results, validated on the OpenWebText using five highly customized filter tasks, provide strong evidence of SIEVE's capability to match GPT-4o's accuracy while significantly reducing computational expenses. The experiments on Datacomp-LM also demonstrates SIEVE's potential in filtering general domain pretraining data.

While our initial study presents promising results, we acknowledge that there are several avenues for future enhancement and exploration. For future work, one can scale SIEVE for larger scales of Datacomp-LM challenges. The modular nature of SIEVE also allows for the integration of more advanced active learning algorithms. Additionally, SIEVE serves as an excellent testbed for these algorithms, offering immediate real-world impact. Future work could also investigate the incorporation of semi-supervised learning techniques to further reduce annotation costs following the framework proposed by~\citet{zhang2024labelbench}. Moreover, while our current implementation focuses on T5 and DeBERTa architectures, future research could examine the efficacy of SIEVE with a broader range of pretrained model architectures for transfer learning. Lastly, exploring the use of more powerful models beyond GPT-4o, such as o1, for handling complex filtering prompts could extend the capabilities of SIEVE to even more challenging scenarios.


\subsubsection*{Acknowledgments}
This work is supported by NSF AI-EDGE institute (NSF Award 2112471). We would like to thank Siddharth Suresh for his help with hosting the human study. We also thank Lalit Jain and Andrew Wagenmaker for providing insightful feedback on our manuscript.

\clearpage

\bibliography{reference}
\bibliographystyle{icml2025}
%%%%%%%%%%%%%%%%%%%%%%%%%%%%%%%%%%%%%%%%%%%%%%%%%%%%%%%%%%%%

\newpage

\onecolumn
\appendix

\section{Extended Literature Review} \label{apx:lit}
\paragraph{Active Learning}
As part of our algorithm, we use an agnostic active learning algorithm for identifying the TRM threshold. Agnostic active learning has been widely studied in the classical PAC learning setups, where the labels of any particular example is inherently noisy. Our procedure is a direct application of \citet{jamieson2022interactive}, which was inspired by \citet{dasgupta2007general}. The algorithm is proven to be near minimax optimal in these literature. In addition, our algorithm can also be seen as an instance of the algorithm proposed by~\citet{katz2021improved} for threshold classifiers, where they also prove such algorithm is near instance-optimal. In this paper, we also prove the first bound towards balancedness of labeled examples in agnostic active learning, focusing on the class of threshold classifiers.

\paragraph{Knowledge Distillation}
Knowledge distillation is a technique where a smaller ``student" model learns to emulate a larger, more sophisticated ``teacher" model. With the increasing capabilities of large language models (LLMs) across diverse tasks, recent research has explored using LLMs as annotators to train domain-specific student models. For a comprehensive review, see \citet{tan2024large}. While most research in LLM knowledge distillation focuses on knowledge extraction methods, few studies address the high computational cost of using LLMs for large-scale annotation. Recent work by \citet{zhang2023llmaaa} and \citet{rouzegar2024enhancing} has begun to tackle this issue by using active learning.
Our paper applies knowledge distillation to the specific problem of data filtering. We employ a straightforward approach, using LLM annotations as binary classification labels to actively fine-tune an encoder-based model. Future research could explore using GPT-4's chain-of-thought outputs to distill a decoder-based student model within the multi-task learning framework proposed by \citet{hsieh2023distilling}.
It's worth noting that classic knowledge distillation work, such as \citet{hinton2015distilling}, trains student classifiers to match the teacher models' output logits. However, in our case, using chain-of-thought filtering prompts makes it impractical to obtain probabilities for the binary decision.

\section{Dataset Details}
\begin{table}[h!]
    \centering
    \begin{tabular}{cc}
        \toprule
        Filter & \shortstack{Imbalance Ratio $\lambda$} \\
        \midrule
        Politics & 0.153\\
        Climate & 0.043\\
        AI & 0.026\\
        \shortstack[c]{Mainstream} & 0.208\\
        Quality & 0.457\\
        \bottomrule
    \end{tabular}
    \caption{Imbalance ratio of minority vs majority decisions, calculated based on $5000$ randomly sampled snippets. $\lambda = \text{\#Minority} /\, \text{\#Majority}$.}
    \label{tab:imbalance}
    \vspace{-\intextsep}
\end{table}

\section{Filtering Prompts} \label{apx:prompts}
\subsection{Politics}
Please analyze the following text snippet and determine if it is relevant to aspects of a presidential election. The snippet may be arbitrarily cut off from a longer article, so please ignore any oddities caused by the truncation and focus on the overall relevance to presidential elections.

\begin{enumerate}
    \item Read the text snippet carefully.
    \item Identify any key terms, phrases, or concepts related to presidential election. These include by are not limited to Candidate information (biographies, backgrounds, policy positions)
    \begin{itemize}
    \item Economic policies and their potential impacts
    \item Social issues and proposed solutions
    \item Foreign policy stances and international relations
    \item Campaign events, debates, and public appearances
    \item Polling data, electoral projections, and voter demographics
    \item Media coverage, endorsements, and fact-checking
    \item Campaign finance and fundraising efforts
    \item Party dynamics and internal politics
    \item Electoral processes, including voting systems and potential reforms
    \item Controversies or scandals involving candidates or their campaigns
    \item Vice presidential candidates and potential cabinet members
    \item Analysis of key battleground states or regions
    \item Digital campaigning strategies and social media presence
    \item Grassroots organizing and volunteer efforts
    \item External events or crises that may influence the election
    \end{itemize}
    
    \item Ignore any abrupt beginning or ending of the snippet and focus on the main content.
    \item Assign either "PASS" for content relevant to presidential election and "FAIL" for those that are irrelevant.
\end{enumerate}

Based on your analysis, determine if the snippet is relevant or irrelevant to the general knowledge of presidential elections. Think step by step and provide your final answer as either "PASS" or "FAIL" at the end of your response and nothing else.

Text snippet: <Insert Text Snippet>

\subsection{Climate}
Please analyze the following text snippet and determine if it is relevant to the general knowledge of climate change. The snippet may be arbitrarily cut off from a longer article, so please ignore any oddities caused by the truncation and focus on the overall relevance to climate change.
\begin{enumerate}
    \item Read the text snippet carefully.
    \item Identify any key terms, phrases, or concepts related to climate change, such as global warming, carbon emission, energy, technology innovation, agriculture, natural resources, pollution, landfill, chemistry, rising sea levels, extreme weather events, climate policies, environmental justice and sustainability.
    \item Assess whether the snippet discusses causes, effects, or solutions to climate change, or provides information that contributes to the understanding of climate change.
    \item Ignore any abrupt beginning or ending of the snippet and focus on the main content.
    \item Assign either "PASS" for content relevant to climate change and "FAIL" for those that are irrelevant.
\end{enumerate}
Based on your analysis, determine if the snippet is relevant or irrelevant to the general knowledge of climate change. Think step by step and provide your final answer as either "PASS" or "FAIL" at the end of your response and nothing else.

Text snippet: <Insert Text Snippet>

\subsection{AI}
Please analyze the following text snippet and determine if it is relevant to aspects of AI. The snippet may be arbitrarily cut off from a longer article, so please ignore any oddities caused by the truncation and focus on the overall relevance to artificial intelligence.
\begin{enumerate}
    \item Read the text snippet carefully.
    \item Find content related to artificial intelligence that discusses computer systems or software designed to perform tasks typically requiring human intelligence, such as visual perception, speech recognition, decision-making, and language translation. Look for explanations of technologies enabling machines to learn from experience, adjust to new inputs, and perform human-like tasks without explicit programming. Include descriptions of systems that can interpret visual information from the world, as well as content about software capable of processing, analyzing, generating, or understanding human language. Seek information on machines or programs that can improve their performance through experience or data. Include discussions of AI applications in various fields such as healthcare, finance, transportation, education, or entertainment. Consider content addressing ethical considerations surrounding AI, including bias, privacy, job displacement, or long-term implications of advanced AI systems. Look for historical accounts of AI development, major milestones, breakthroughs, or setbacks in the field. Include explanations of AI algorithms, their workings, strengths, limitations, and potential applications. Capture debates or discussions about the future of AI, including topics like artificial general intelligence, superintelligence, or potential societal impacts of widespread AI adoption. Include reports on current research, new methodologies, experimental results, or theoretical advancements in AI. Consider content about prominent figures, organizations, or companies significantly contributing to AI research and development. Look for discussions of AI policy, regulation, or governance at organizational, national, or international levels. Include explanations of the relationship between AI and other fields such as robotics, Internet of Things, big data, or quantum computing. Finally, capture content addressing challenges in AI development, such as data quality, computational requirements, or the need for explainable AI systems.

    \item Ignore any abrupt beginning or ending of the snippet and focus on the main content.
    \item Assign either "PASS" for content relevant to AI and "FAIL" for those that are irrelevant.
\end{enumerate}

Based on your analysis, determine if the snippet is relevant or irrelevant to aspects of artificial intelligence, its development, applications, or implications. Think step by step and provide your final answer as either "PASS" or "FAIL" at the end of your response and nothing else.

Text snippet: <Insert Text Snippet>

\subsection{Mainstream Knowledge}
\textbf{We asked for obscure knowledge instead, so any snippet that "Failed" would be considered mainstream knowledge.}

Please analyze the following text snippet and determine if it contains obscure or niche knowledge that less than 10000 people know and understand. The snippet may be arbitrarily cut off from a longer article, so please ignore any oddities caused by the truncation and focus on the overall relevance to artificial intelligence.
\begin{enumerate}
    \item Read the text snippet carefully.
    \item Filter the dataset for content related to obscure, specialized, or highly niche knowledge that is not commonly known or easily accessible to the general public. Include information on rare historical events, obscure scientific theories, uncommon philosophical concepts, extinct languages, highly specialized mathematics, niche literary works, rare medical conditions, uncommon species, esoteric subcultures, mystical practices, experimental technologies, obscure laws, rare geological formations, lesser-known art movements, specialized crafting techniques, rare musical instruments, uncommon culinary practices, obscure sports, theoretical cosmological concepts, and highly specialized areas of archaeology or anthropology. Exclude any content that is commonly known, part of standard education, regularly discussed in popular media, or widely understood by the general public. The ideal content should require specialized knowledge, extensive research, or access to uncommon sources of information, rather than being something an average person would encounter in daily life or through casual exposure to media and education.
    \item Ignore any abrupt beginning or ending of the snippet and focus on the main content.
    \item Assign either "PASS" for obscure and niche knowledge and "FAIL" for those that are common knowledge.
\end{enumerate}

Think step by step. Then, you must provide your final answer as either "PASS" or "FAIL" at the end of your response and nothing else.

Text snippet: <Insert Text Snippet>

\subsection{Quality}

Please analyze the following text snippet and determine if it is high quality or low quality training data for a large language model. The snippet may be cut off abruptly from a longer piece of text, but focus your analysis on the quality factors present in the provided text rather than the awkward truncation.
Quality factors to consider include:
\begin{enumerate}
    \item Evaluate the spelling, grammar, and overall writing quality of the snippet. Note any errors or inconsistencies that could negatively impact the model's learning.
    \item Assess the factual accuracy and reliability of the information presented in the snippet. Consider whether the content appears trustworthy and well-researched.
    \item Analyze the clarity, coherence, and logical flow of ideas in the snippet. Determine if the text is easy to understand and follow.
    \item Gauge the breadth and depth of knowledge conveyed in the snippet. Consider whether the content provides valuable information or insights on the topic at hand.
    \item Examine the neutrality and objectivity of the tone and perspective presented in the snippet. Consider if the text appears biased or presents a balanced viewpoint.
    \item Based on the above factors, determine if the snippet is:\\
    PASS: High quality training data\\
    FAIL: Low quality training data
\end{enumerate}
Think step by step and answer with either PASS or FAIL as your final decision in the end and nothing else.

Text snippet: <Insert Text Snippet>

\newpage
\section{Analysis Proof} \label{apx:proof}
\begin{proof}
    Let $A = \sum_{j: f(x_j)\in[\underline{\mu},\bar{\mu}]}\eta_{j}$ and $B=\sum_{j: f(x_j)\in[\underline{\mu},\bar{\mu}]}1 - 2\eta_{j}$, we can rewrite the imbalance ratio into $\lambda(\underline{\mu},\bar{\mu}) = \frac{A}{A + B}$. Since $N(\gamma_1 - \gamma_0) = \sum_{j:f(x_j)\in (\underline{\mu}, \bar{\mu}]} 1 - 2\eta_j \geq B - LN$, we can lower bound the balancedness by $\frac{A}{A + N(\gamma_1 - \gamma_0) + LN}$.

    As the lower bound $\frac{A}{A + N(\gamma_1 - \gamma_0) + LN}$ increases as $A$ increases, we would now like to prove a lower bound of $A=\sum_{j: f(x_j)\in[\underline{\mu},\bar{\mu}]}\eta_{j}$.

    Recall $\pi(1), ..., \pi(N)$ is the ordering of examples based on sigmoid score. We let $\pi^{-1}(\cdot)$ denote the inverse mapping of $\pi$, so that $\pi^{-1}(\pi(i)) = i$. We let $\underline{r} = \min(\{j: f(x_{\pi(j)}) \in [\underline{\mu}, s^\star]\})$, $\bar{r} = \max\{j: f(x_{\pi(j)})\in (s^\star, \bar{\mu}]\}$ and $r^*$ denote the index where $f(x_{\pi(r^\star)}) = s^\star$.

    First note since class 0 is the minority class, we must have $r^\star \neq N$.
    Since $r^\star \neq 0$ and $r^\star \neq N$, we must have $\eta_{\pi(r^\star)} \geq 0.5$ and $\eta_{\pi(r^\star+1)} \leq 0.5$. Otherwise, $r^\star + 1$ or $r^\star - 1$ will have lower risk than $R(\eta_{\pi(r^\star)})$. By the smoothness definition above, we further have $\forall j \leq r^\star, 0.5 - (r^\star - j)L \leq \eta_{\pi(j)} \leq 0.5 + (r^\star - j + 1)L$, and  $\forall j \geq r^\star, \eta_{\pi(j)} \geq 0.5 - (j - r^\star + 1)L$.
    
    $A=\sum_{j: f(x_j)\in[\underline{\mu},\bar{\mu}]}\eta_{j}$ can then be rewritten in the ranked format as $A = \sum_{j \in [\underline{r}, \bar{r}]} \eta_j$.

    First, we divide the sampling range into $[\underline{r}, r^\star]$ and $[r^\star + 1, \bar{r}]$. When sampling in $[\underline{r}, r^\star]$, we have
    \begin{align} \label{eqn:risk_bound_left}
        & R(f(x_{\pi(\underline{r})})) - R(s^\star) = \gamma_0 > 0 \implies \nonumber\\
        &N\gamma_0 = \sum_{j\in(\underline{r}, r^\star]} \eta_{(j)} - \sum_{j\in(\underline{r}, r^\star]} (1 - \eta_{(j)}) \geq -1 + \sum_{j\in[\underline{r}, r^\star]} \eta_{(j)} - \sum_{j\in[\underline{r}, r^\star]} (1 - \eta_{(j)}).
    \end{align}

    When sampling in $[r^\star + 1, \bar{r}]$, since $R(\bar{r}) - R(r^\star) = \gamma_1$, we must have
    \begin{align} \label{eqn:risk_bound_right}
        N\gamma_1 = N\cdot (R(\bar{r}) - R(r^\star))) = \sum_{j\in(r^\star, \bar{r}]} 1 - 2\eta_{(j)} = \sum_{j\in[r^\star + 1, \bar{r}]} (1 - \eta_{(j)}) - \sum_{j\in[r^\star + 1, \bar{r}]} \eta_{(j)}.
    \end{align}
    
    % Together, we can bound the balancedness by
    % \begin{align*}
    %     & \frac{\sum_{j\in[\underline{r},\bar{r}]}\eta_{(j)}}{\sum_{j\in[\underline{r},\bar{r}]}(1-\eta_{(j)})} = \frac{\sum_{j\in[\underline{r}, r^\star]} \eta_{(j)} + \sum_{j\in[r^\star + 1, \bar{r}]} \eta_{(j)}}{\sum_{j\in[\underline{r}, r^\star]} (1 - \eta_{(j)}) + \sum_{j\in[r^\star + 1, \bar{r}]} (1 - \eta_{(j)})} \\
    %     &\geq \frac{\sum_{j\in[\underline{r}, r^\star]} \eta_{(j)} + \sum_{j\in[r^\star + 1, \bar{r}]} \eta_{(j)}}{(\sum_{j\in[\underline{r}, r^\star]} \eta_{(j)}) - N\gamma_0 + 1 +(\sum_{j\in[r^\star + 1, \bar{r}]} \eta) + N\gamma_1} \\
    %     &= \frac{\sum_{j\in[\underline{r}, \bar{r}]} \eta_{(j)}}{\sum_{j\in[\underline{r}, \bar{r}]} \eta_{(j)} + N (\gamma_1 - \gamma_0) + 1}
    % \end{align*}

   To obtain the lower bound of $\sum_{j\in[\underline{r}, \bar{r}]} \eta_{(j)}$, we start by bounding $\bar{r}$ and $\underline{r}$. Specifically, by \eqref{eqn:risk_bound_right}, we have
   \begin{align*}
       N\gamma_1 & = \sum_{j\in[r^\star + 1, \bar{r}]} (1 - 2\eta_{(j)})  \leq \sum_{j\in[r^\star + 1, \bar{r}]} (1 - 2(0.5 - (j - r^\star)L)) \\
       &= \sum_{j\in[r^\star + 1, \bar{r}]} 2(j - r^\star)L = \sum_{j = 1}^{\bar{r} - r^\star} j L = (\bar{r} - r^\star)(\bar{r} - r^\star + 1)L \leq L (\bar{r} - r^\star + 1)^2.
   \end{align*}
   As a result $\bar{r} \geq \sqrt{\frac{N\gamma_1}{L}} + r^\star - 1$. Let $\alpha_1 = \sqrt{\frac{N\gamma_1}{L}} - 1$, we then have $\bar{r} \geq r^\star + \alpha_1$.

   Similarly, we have
   \begin{align*}
       & N\gamma_0 \leq \sum_{j \in [\underline{r}, r^\star]}(2\eta_{(j)} - 1) \leq \sum_{j \in [\underline{r}, r^\star]}(2\cdot (0.5 + (r^\star - j + 1)L) - 1) \\
       &= \sum_{j \in [\underline{r}, r^\star]}2(r^\star - j + 1)L = \sum_{j=1}^{r^\star - \underline{r} + 1} 2jL \leq (r^\star - \underline{r} + 2)^2 L,
   \end{align*}
   so $\underline{r} \leq r^\star -\sqrt{\frac{N\gamma_0}{L}} + 2$. Let $\alpha_0 := \sqrt{\frac{N\gamma_0}{L}} - 2$, we then have $\underline{r} \leq r^\star - \alpha_0$

   Now, we can bound $\sum_{j\in[r^\star + 1, \bar{r}]} \eta_{(j)}$ by the following
   \begin{align*}
       & \sum_{j\in[r^\star + 1, \bar{r}]} \eta_{(j)} = \sum_{j=r^\star + 1}^{r^\star + \alpha_1} \eta_{(j)} \geq \sum_{j=r^\star + 1}^{r^\star + \alpha_1} 0.5 - (j - r^\star)L \\
       & = \sum_{j=1}^{r^\star + \alpha_1} 0.5 - jL = \frac{\alpha_1}{2} - \frac{\alpha_1(\alpha_1 + 1)L}{2} \\
       &= \frac{(1 - L)\alpha_1 - \alpha_1^2L}{2}.
   \end{align*}

   Similarly, we can bound $\sum_{j\in[\underline{r}, r^\star]} \eta_{(j)}$ by the following
   \begin{align*}
       &\sum_{j\in[\underline{r}, r^\star]} \eta_{(j)} = \sum_{j=r^\star - \alpha_0}^{r^\star} \eta_{(j)}
       \geq \sum_{j=r^\star - \alpha_0}^{r^\star} 0.5 - (r^\star - j)L \\
       &=\sum_{j=0}^{\alpha_0} 0.5 - jL = \frac{\alpha_0 + 1}{2} - \frac{\alpha_0(\alpha_0 + 1)L}{2} \\
       & = \frac{(1-L)\alpha_0 - \alpha_0^2L + 1}{2}
   \end{align*}
    Together, we have 
    \begin{align*}
    &\sum_{j\in[\underline{r}, \bar{r}]} \eta_{(j)} \geq \frac{(1-L)(\alpha_0 + \alpha_1) - (\alpha_0^2 + \alpha_1^2)L + 1}{2}. \\
    &\geq \frac{1}{2}((1-L)(\alpha_0 + \alpha_1) + 1) \geq \frac{1}{2}(2(1-L)\sqrt{\frac{N\underline{\gamma}}{L}} -2) = (1-L)\sqrt{\frac{N\underline{\gamma}}{L}} - 1
    % &= \frac{1}{2}\left((1-L)(\sqrt{\frac{N\gamma_1}{L}} + \sqrt{\frac{N\underline{\gamma}}{L}} - 3) -  + 1\right)
    \end{align*}.

    For the edge case, since the interval must have more than three snippets, we can bound $\sum_{j\in[\underline{r}, \bar{r}]} \eta_{(j)} \geq 1.5 - 2L$. Therefore, we have $\sum_{j\in[\underline{r}, \bar{r}]} \eta_{(j)} \geq \max(1.5 - 2L, (1-L)\sqrt{\frac{N\underline{\gamma}}{L}} - 1)$.

   Finally, we can bound the balancedness by
   \begin{align*}
   &\frac{\sum_{j\in[\underline{r},\bar{r}]}\eta_{(j)}}{\sum_{j\in[\underline{r},\bar{r}]}(1-\eta_{(j)})} \geq \frac{\sum_{j\in[\underline{r}, \bar{r}]} \eta_{(j)}}{\sum_{j\in[\underline{r}, \bar{r}]} \eta_{(j)} + N (\gamma_1 - \gamma_0) + LN} \\
   &\geq 1 - \frac{N (\gamma_1 - \gamma_0) + LN}{\sum_{j\in[\underline{r}, \bar{r}]} \eta_{(j)} + N (\gamma_1 - \gamma_0) + LN} \\
   &\geq 1 - \frac{N \bar{\gamma} + LN}{\sum_{j\in[\underline{r}, \bar{r}]} \eta_{(j)} + N \bar{\gamma} + LN} \\
   & \geq 1 - \min(\frac{N\bar{\gamma} + LN}{1.5 + N\bar{\gamma} + LN - 2L}, \frac{N\bar{\gamma} + LN}{(1-L)\sqrt{\frac{N\underline{\gamma}}{L}} + N\bar{\gamma} + LN - 1}) \\
   & \geq 1 - \min(\frac{N\bar{\gamma} + LN}{1.5 - 2L}, \sqrt{L}\cdot\frac{N\bar{\gamma} + LN + 1}{(1-L)\sqrt{N\underline{\gamma}}})
   \end{align*}

\end{proof}

\section{Training Details} \label{apx:training}
Our model is fine-tuned using the AdamW optimizer with a cosine learning rate schedule. For every training of $f$ in Algorithm~\ref{alg:active}, we train up to $5$ epochs. When measuring performance, we use a separate validation set to find the highest performance checkpoint. For focal loss, we use $\gamma = 5$, and $\alpha$ is set to the imbalance ratio estimated from Table~\ref{tab:imbalance} for the minority class.

\section{Cost Comparisons} \label{ssec:cost}
\begin{table*}[t]
    \centering
    \scalebox{0.8}{
    \begin{tabular}{llcccccc}
    \toprule
    Filter & Method &  \shortstack{\#Queries to \\ GPT-4o} & \shortstack{Bal. Accuracy\\(GPT-4o as GT)} & \shortstack{GPT-4o\\Cost} & \shortstack{Lightweight \\Model Training} & \shortstack{Lightweight \\Model Inference} & \shortstack{Total\\Cost}
    \\
    \midrule
    \multirow{3}{*}{Climate} & Random & 25K & \textbf{96.6\%} & \$125 & \$20 & $\$50$ & \$195\\
    & GPT-4o & 13.5M & \textbf{96.6\%} & \$67,000 & \$0 & \$0 & \$67,000\\
    & SIEVE (Ours) &  \textbf{7.5K} & \textbf{96.7\%} & \$40 & \$30 & \$50 & \textbf{\$120}\\
    \midrule
    \multirow{3}{*}{AI} & Random & 25K & \textbf{95.7\%} & \$125 & \$20 & $\$50$ & \$195\\
    & GPT-4o & 13.5M & \textbf{95.5\%} & \$67,000 & \$0 & \$0 & \$67,000\\
    & SIEVE (Ours) &  \textbf{6K} & \textbf{95.7\%} & \$30 & \$30 & \$50 & \textbf{\$110}\\

    \midrule
    \multirow{4}{*}{Politics} & Random & 100K & 95\% & \$500 & \$20 & $\$50$ & \$570\\
    & SIEVE (Ours) & \textbf{25K} & 95\% & \$125 & \$30 & \$50 & \$205\\
    & GPT-4o & 13.5M & \textbf{95.6\%} & \$67,000 & \$0 & \$0 & \$67,000\\
    & SIEVE (Ours) & \textbf{60K} & \textbf{95.6\%} & \$300 & \$120 & \$50 & \textbf{\$470}\\
    \bottomrule
    \end{tabular}
    }
    \caption{Cost breakdown of climate, AI and climate filters. Total cost is consisted of GPT-4o querying cost, lightweight model training inference costs.}
    \label{tab:cost_comparison}
\end{table*}
Table~\ref{tab:cost_comparison} presents a comprehensive breakdown of SIEVE's computational costs, consisting of GPT-4o query costs, lightweight model training costs, and inference costs for filtering the entire OpenWebText dataset. The sum of GPT-4o query and training costs represents the total model distillation expense for SIEVE. Our experiments indicate that querying GPT-4o for 1000 snippets costs approximately \$5. During lightweight model inference, we utilize low-precision inferences to significantly reduce computational cost. The lightweight model training cost includes costs for model fine-tuning at each iteration of Algorithm \ref{alg:active}, inference costs for computing sigmoid scores of data in stream, and negligible uncertainty update expenses. We calculated these costs based on the hourly rate for 8$\times$H100 GPUs at \$16 (pricing by many smaller cloud providers), multiplied by the actual time spent on each experiment. Together, these components provide a detailed overview of the computational resources required for SIEVE's implementation and application to the entire dataset.

\paragraph{Stream-Based vs Pool-Based Active Distillation.}
The choice of a stream-based active learning approach for SIEVE is justified by its significant cost advantages over pool-based methods. While stream-based algorithms process snippets only once, pool-based active learning requires repeated forward inference on the entire dataset of 13.5M snippets for each batch of queried snippets. This difference translates to substantial additional costs: approximately \$1800 for the politics filter and \$750 for the climate filter. These extra expenses would dramatically increase the overall cost of active distillation. Our decision to develop a stream-based active learning algorithm for SIEVE, particularly effective in imbalanced scenarios, is thus strongly supported by these cost considerations, ensuring a more economically viable solution for our system.

\section{Ablation on Model Choice} \label{apx:model_ablation}
\begin{table}[h]
    \centering
    \scalebox{0.8}{
    \begin{tabular}{llc}
    \toprule
    Filter & Model &  \shortstack{\#Queries to reach GPT-4o\\ performance within 0.3\%}\\
    \midrule
    \multirow{2}{*}{Climate} & T5-large & 7,500\\
     & DeBERTa-v2-xlarge & 9,000\\
    \midrule
    \multirow{2}{*}{AI} & T5-large & 6,000\\
    & DeBERTa-v2-xlarge &  9,000\\
    \bottomrule
    \end{tabular}
    }
    \caption{Ablation study of model choice. We see that DeBERTa and T5 classifiers are both capable of reaching similar performance as GPT-4o within a reasonable budget. Here, the active learning algorithm has been fixed to use Algorithms~\ref{alg:active} for SIEVE.}
    \label{tab:model_ablation}
\end{table}
\end{document}

%% file: math_commands.tex
\usepackage{amsmath,amsfonts,bm}

\def\eqref#1{equation~\ref{#1}}
\def\1{\bm{1}}

\DeclareMathAlphabet{\mathsfit}{\encodingdefault}{\sfdefault}{m}{sl}
\SetMathAlphabet{\mathsfit}{bold}{\encodingdefault}{\sfdefault}{bx}{n}

\DeclareMathOperator*{\argmin}{arg\,min}